\pagebreak

\documentclass[preprint]{elsarticle}

\usepackage{lineno}

\usepackage{graphicx}
\usepackage[caption=false, font=normalsize, labelfont=sf, textfont=sf]{subfig}

\usepackage{tabularx}
\usepackage{nameref}
\usepackage{makecell}

\usepackage{gensymb}

\usepackage{tikz} 
\newcommand{\orcidicon}{\includegraphics[width=0.32cm]{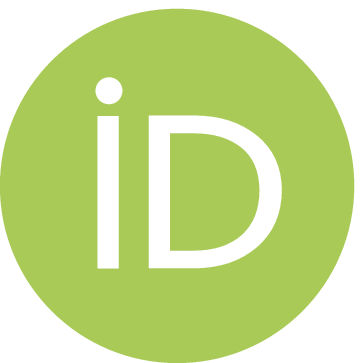}}
\foreach \x in {A, ..., Z}{%
\expandafter\xdef\csname orcid\x\endcsname{\noexpand\href{https://orcid.org/\csname orcidauthor\x\endcsname}{\noexpand\orcidicon}}
}

\usepackage{hyperref}

\modulolinenumbers[5]

\journal{SLAS Technology} 

\begin{document}

\begin{frontmatter}

\title{Towards Robotic Laboratory Automation Plug \& Play: \\ Survey and Concept Proposal on Teaching-free Robot Integration with the LAPP Digital Twin}

%

\author[adamadress,adamsecondaryadress]{Ádám Wolf \orcidA{}\corref{mycorrespondingauthor}}
\cortext[mycorrespondingauthor]{Corresponding author}
\ead{adam.wolf@takeda.com}
\address[adamadress]{Takeda Manufacturing Austria AG, Industriestraße 67, A-1221 Wien, Austria}
\address[adamsecondaryadress]{Doctoral School of Applied Informatics and Applied Mathematics, Óbuda University}

\author[stefanadress]{Stefan Romeder-Finger \orcidD{}}
\address[stefanadress]{Baxalta Innovations GmbH, a Takeda company}

\author[karolyadress]{Károly Széll \orcidC{}}
\address[karolyadress]{Alba Regia Technical Faculty, {\'O}buda University, H-8000 Sz{\'e}kesfeh{\'e}rv{\'a}r, Hungary}
\author[peteradress]{Péter Galambos \orcidB{}}
\address[peteradress]{Antal Bejczy Center for Intelligent Robotics, Óbuda University}

\begin{abstract}
The Laboratory Automation Plug \& Play (LAPP) framework is an over-arching reference architecture concept for the integration of robots in life science laboratories. The plug \& play nature lies in the fact that manual configuration is not required, including the teaching of the robots. In this paper a digital twin (DT) based concept is proposed that outlines the types of information that must be provided for each relevant component of the system. In particular, for the devices interfacing with the robot, the robot positions must be defined beforehand in a device-attached coordinate system (CS) by the vendor. This CS must be detectable by the vision system of the robot by means of optical markers placed on the front side of the device. With that, the robot is capable of tending the machine by performing the pick-and-place type transportation of standard sample carriers. This basic use case is the primary scope of the LAPP-DT framework. The hardware scope is limited to simple benchtop and mobile manipulators with parallel grippers at this stage. This paper first provides an overview of relevant literature and state-of-the-art solutions, after which it outlines the framework on the conceptual level, followed by the specification of the relevant DT parameters for the robot, for the devices and for the facility. Finally, appropriate technologies and strategies are identified for the implementation.
\end{abstract}


\end{frontmatter}


\section*{Introduction}
This article is the second in the series \emph{Towards Robotic Laboratory Automation Plug \& Play}. The first paper outlined the concept in a high-level fashion and stated the fundamental goal for the Laboratory Automation Plug \& Play (LAPP) framework: to provide a comprehensive, over-arching reference architecture and abstraction layer for robot-focused laboratory automation \cite{Wolf2021TowardsFramework}. The top level in this context is the process representation and scheduling layer, which serves the orchestration of the laboratory workflows. This includes communication and control aspects considering various types of (semi-)automated laboratory devices (such as liquid handlers, plate storage, centrifuges, incubators and readers) as well as laboratory robots. 

The framework aims to simplify the integration and set-up process for these types of robots by freeing the integrator/user from the burden of manually setting up the communication and teaching the robot motions. As a crucial part of the LAPP framework, the digital twin (DT) layer enables the storing and sharing of the robot-relevant information in regard to the various components of the system. Most importantly, the positions for  arm control must be stored in a standardized fashion (encoded in the DT) and provided with the device out-of-the-box. Using these pieces of information, a mobile manipulator (MoMa) will be capable — after localizing the device with a fiducial marker — to tend the device in a plug \& play manner.

The LAPP framework focuses at first on a specific type of MoMa that is now used by multiple solution providers for sample transportation in automated laboratories \cite{Kleine-Wechelmann2022DesigningLaboratoryb, BioseroLab, RoboticProjects}. For this, the typical lab automation MoMa anatomy is considered. Such a robot consists of the following components:
\begin{itemize}
    \item A wheeled autonomous mobile base capable of simultaneous localization and mapping (SLAM)
    \item A robot arm of four to seven degrees of freedom (DoF)
    \item A vision system that aids the fine position detection for gripping and manipulation planning
\end{itemize}
A MoMa can be considered as the most complex robot system in our scope, which represents the broadest variety of components. As such, simpler robots (e.g., stationary or rail-mounted arms) can be derived from this set by omitting certain components (e.g., mobile navigation). On the use case side, the concept is developed at the first stage for simple pick-and-place type sample transportation and machine tending without any physical interaction with the equipment. Complex robot-equipment interactions will be elaborated in follow-up articles.


The paper starts with the \nameref{sec:methods} section, where the strategy for elaborating the concept is outlined. Then a section on \nameref{sec:literature} provides an overview of the relevant solutions in regard to the vertical of robot-centered laboratory automation systems. In the section on \nameref{sec:man_teaching}, the lowest level is addressed by reviewing how prevalent manual teaching approaches work. To conclude the literature and state-of-the-art review, the digital twin (DT) approach is reviewed, which will serve as the basis of the LAPP-DT proposal. In \nameref{sec:teaching_free} the concept is proposed, followed by identifying and discussing \nameref{sec:challenges}. After that, key \nameref{sec:technologies} are discussed and the pieces of the technological stack are outlined. Finally, \nameref{sec:future_work} and next steps are introduced, and a \nameref{sec:conclusion} is provided.

\section*{Methods} \label{sec:methods}
The LAPP concept aims to provide an overview of the system components and their interfaces, outlining a comprehensive integration framework as a reference architecture model. One crucial aspect of this is the information representation layers (or digital twins), which can facilitate teaching-free robot integration. Both the over-arching LAPP framework and the DT layer (subject of this paper) are created and elaborated in a scalable and extendable manner throughout an evolutionary development process. This development approach can be outlined with the following bullet points:
\begin{itemize}
    \item The concept is first formulated in an abstract and high-level fashion, as published in \cite{Wolf2021TowardsFramework}. It will then be concretized and adapted with the involvement of the professional community throughout continuous evolution.
    \item The definition of the framework fundamentally focuses on the already existing and established building-block technologies but will be suitable to incorporate new developments, such as advances in collaborative robotics.
    \item Initially, a limited set of typical laboratory tasks that are already subject to a certain level of automation/robotization will be considered. Later, as the point above suggests, new technologies will enable covering more complex tasks that the framework also must be able to incorporate.
\end{itemize}

\section*{Literature and the state-of-the-art} \label{sec:literature}

\subsection*{Manual teaching of laboratory robots} \label{sec:man_teaching}
\begin{figure}
    \centering
  \subfloat[Benchtop robot\label{fig:cs_benchtop}]{%
       \includegraphics[width=0.4\linewidth]{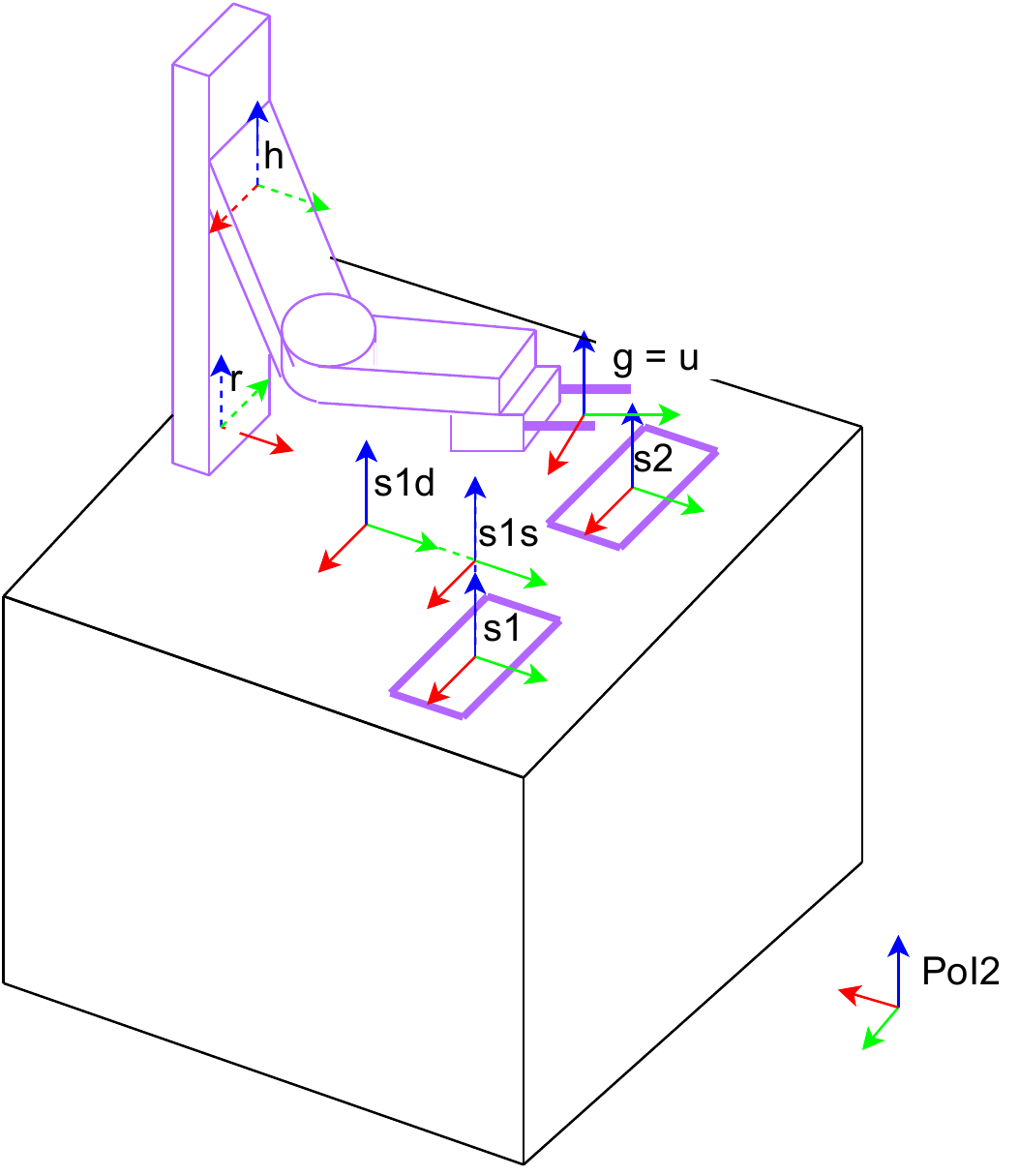}}
    \hspace{5mm}
  \subfloat[Mobile robot\label{fig:cs_mobile}]{%
        \includegraphics[width=0.7\linewidth]{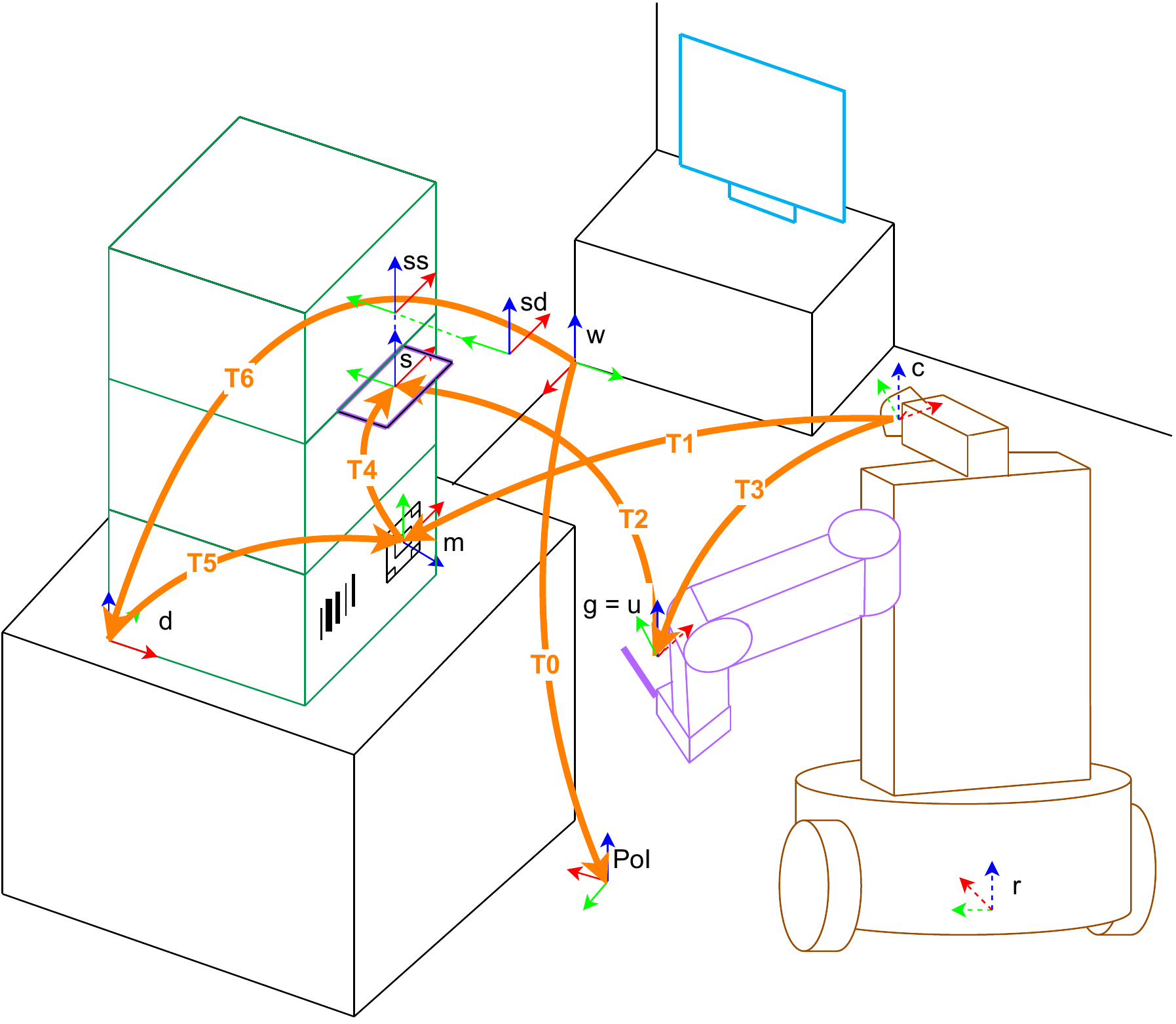}}
  \caption{Coordinate systems in the LAPP environment, d: device CS, g: robot's tool center point (TCP), h: home position for robot  m: fiducial marker for device, r: robot's base, PoI: base pose for station, s: hand-over site (plate nest) of device, sd: device approach position, ss: site approach position, T0...T6: relational frame definitions (T0: world-to-PoI, T1: camera-to-marker, T2: TCP-to-site, T3: camera-to-TCP, T4: marker-to-site, T5: device-to-marker, T6: world-to-device), u: stand-by position for robot, w: world. \\ Lines: Orange: live robot-level, Green: stored in the LAPP-DT, Dashed: inaccurate transformation (originating from base odometry), Solid: accurate transformation (originating from robot kinematics, marker detection or DT-stored positions) }
  \label{fig:coord_sys} 
\end{figure}

At the current state of technology, robotic manipulators in laboratory automation are mostly intended for transporting standard ANSI/SLAS-conform\footnote{Meets the Standards ANSI/SLAS 1-2004 through ANSI/SLAS 4-2004} microplates between different laboratory devices. To make this possible, in most cases, online teaching takes place when setting up the robotic system. This means that the positions are manually set either by moving the robot by hand or by jogging it with the controller. Figure \ref{fig:coord_sys} shows the coordinate systems (frames) and positions that are typically used in laboratory robotics scenarios.

 \pagebreak
 
In the case of stationary robots, either the robot configuration is stored for each position in the form of joint values, or the position is directly prescribed in relation to the robot's base CS (r in Fig. \ref{fig:cs_benchtop}). Between these taught positions, different types of movements are possible. Usually a sequence of intermediary approach positions are defined for the robot, before it reaches a final hand-over site with its gripper tool center point (TCP). This is to make sure that no collisions occur between the robot and its surroundings, including the device it is supposed to load with samples. A typical sequence of steps for a pick-and-place task from site 1 (s1) to site 2 (s2) is presented below, using the coordinate systems in Fig. \ref{fig:cs_benchtop}:

\begin{itemize}
    \item The robot starts in its home position (h)
    \item Performs a MoveJ-type movement to "untangle" itself and arrives at the standby configuration at the position (u)
    \item Moves linearly (MoveL) to a device-approach position (s1d)
    \item MoveL to a site-approach position (s1s)
    \item MoveL to the final site hand-over position (s1)
    \item Grips the plate
    \item MoveL back to s1s
    \item MoveL back to s1d
    \item MoveL to s2d (not displayed)
    \item MoveL to s2s (not displayed)
    \item MoveL to s2 (not displayed)
    \item Releases the plate
    \item MoveL back to s2s
    \item MoveL back to s2d
    \item MoveL to u and returns to standby/ready state
\end{itemize}

As described in \cite{Wolf2021TowardsFramework}, in the case of mobile manipulators, on the other hand, the positions are defined in relation to a fiducial marker, which is in most cases an optical augmented reality (AR) marker \cite{Garrido-Jurado2014AutomaticOcclusion} (Fig. \ref{fig:cs_mobile} m). This is necessary because the precision of the mobile base (Fig. \ref{fig:cs_mobile} r) is not sufficient for the positions to be defined in a word-fixed CS (Fig. \ref{fig:cs_mobile} w).

The robot itself is keeping track of its coordinate systems (frames) by maintaining a transfer tree, where relations between the frames are expressed. For example, the pose of the camera frame and the TCP are both known in relation to the robot's base frame. When the arm is moving, the system regularly updates the corresponding chain of frames based on the kinematic model and the joint angles (forward kinematics).

\pagebreak

The teaching process for marker-guided MoMas typically consists of the following steps, as deduced from the feature definitions for Astech Projects' MoMa \cite{Feature_definitions/ch/unitelabsGitLab} and from the Fraunhofer IPA's Kevin Webinar \cite{224YouTube} (Markings of Fig. \ref{fig:cs_mobile} are used.):

\begin{itemize}
    \item The operator drives the robot to the station and makes sure the marker is in the camera's field of view
    \item This pose of the mobile robot is stored as a point of interest (PoI) with the station's name \(T_0(w  \rightarrow PoI)\)
    \item The vision system determines the transformation from the camera frame to the marker frame: \(T_1(c  \rightarrow m)\)
    \item The operator moves the arm to the hand-over position so that the TCP and the site coordinate systems align: \(T_2(g = s)\)
    \item Based on the robot kinematics model, the relation between the camera frame and the TCP ( \(T_3(c \rightarrow g)\) ) is determined
    \item The relation between the marker and the hand-over site is calculated: \( T_4(m \rightarrow s)\)
\end{itemize}

Similar to stationary robots, the intermediary positions (ss and sd in Fig. \ref{fig:cs_mobile}) can be taught additionally. The concept proposed in this paper provides a solution for eliminating the need for manual teaching and setup. For that, the standardized representation of the teaching positions and other relevant parameters based on the digital twin approach is presented in Section \nameref{sec:teaching_free}.

\subsection*{The digital twin approach} \label{sec:dt}
One of the most important aspects of the LAPP framework is that it provides an information layer of the assets of the laboratory. As proposed in \cite{Wolf2021TowardsFramework}, a LAPP-enabled laboratory device has a DT representation containing its parameters. These include both parameters that are constant and assigned when the device is manufactured as well as parameters that change during the operation of the device and that describe the current state. The approach of a virtual representation of physical entities aligns with the digital twin (DT) concept. In this Section, an overview of the DT concept is provided and the relevant nomenclature is specified. 
Originally, this concept was proposed by Michael Grieves at NASA \cite{Grieves2016DigitalSystems} for product life cycle management purposes. Maintaining a virtual representation of a \emph{physical entity} and implementing bi-directional data connections enables a variety of virtual operations ranging from modeling through testing to optimization. All these operations correspond to a specific sub-space of the digital twin. Generally, data are fed from the real space to the virtual space, and information is fed back alongside with processes. A DT accompanies the product throughout its entire life cycle and, according to the specific stage, can take up different forms. In parallel to the design phase of the product, a \emph{prototype} of the DT is also developed, which is then \emph{instantiated} for each physical product individually. The whole set of these instances make up the \emph{aggregate}, and they all reside in the corresponding \emph{environment}. The \emph{state} of both the physical and the virtual entities are stored in the form of \emph{para\-meters}. These can either be static (so-called \emph{prototype}) parameters, which have the same value for each piece of the same product, or they can correspond to a specific physical specimen, in which case they are called \emph{instance} parameters.

Since the proposal of the original concept, the DT notion has been adapted to a broad variety of use-cases. Jones et al. \cite{Jones2020CharacterisingReview} provide a systematic review of the related literature by means of a thematic analysis and characterization. The paper also consolidates the terminology in regard to the terminology that will be followed in the context of the present article. Jones et al. also specify the parameters that are often used for digital twins, which are summarized in Table \ref{tab:dt_params}.

\begin{table}[ht]
\caption{Parameters of the digital twin \cite{Jones2020CharacterisingReview}.}
\label{tab:dt_params}
\begin{tabular}{ p{0.2\linewidth} p{0.7\linewidth} }
\textbf{Parameter}  & \textbf{Characteristics}                                                                                                                            \\ \hline
{Form}          & \makecell[l]{The geometric structure of the entity\\ Can contain dimensions, CS definitions\\ and 3D models} \\ \hline
{Functionality} & \makecell[l]{The entity's capabilities, movement or purpose,\\ Definition of the control interface and parameters}                                     \\ \hline
Health              & State of the entity with respect to its ideal state                                                                                                 \\ \hline
{Location}      & \makecell[l]{Position and orientation (pose)\\ With respect to the environment, layout or other entity}                   \\ \hline
{Process}       & \makecell[l]{Activities in which the entity is engaged,\\ Scheduling parameters and models}                               \\ \hline
Time                & \makecell[l]{Duration to complete an activity,\\ Timestamp}                                                                \\ \hline
State               & Measured (live) values of entity and environment parameters                                                                                     \\ \hline
Performance         & Measured operation with respect to the optimal operation                                                                                            \\ \hline
Environment         & The physical and virtual environment as a parameter                                                                                                 \\ \hline
{Miscellaneous} & \makecell[l]{Requirements\\ Qualification}                                                                                 \\ \hline
\end{tabular}
\end{table}

The DT concept is broadly applied for smart manufacturing use cases \cite{Rosen2015AboutManufacturing, Negri2017ASystems, Lu2020DigitalIssues, Schleich2017ShapingEngineering}, and there are also numerous examples of robotics applications, where the DT may have different goals. These range from the education of future engineers \cite{Verner2018RobotProject, Magrin2021CreatingRobotics} through human-machine interactions \cite{Malik2018DigitalSetting, Pairet2019AInteraction, BaranyiIntroducingI} and control \cite{Vu2021DesignRV-12SD} all the way to  programming \cite{Kuts2020DigitalCase, Burckhardt2018}. Tipary and Erdős propose a design approach that utilizes parametric digital twins of robotic workcells for planning and programming \cite{Erdos2020TransformationTwins, Tipary2021GenericTwin}. To adapt the off-line-created robot program to the physical workcell, calibration and manual adjustments are needed in on-line mode most of the time. To bridge this gap, they define the so-called Digital twin closeness as the measure of the "geometric difference between the digital and physical counterparts of digital twins for robotic workcells". Their DT comprises the models in Table \ref{tab:dt_tipary}. The LAPP-DT concept presented in this paper follows an approach that shares many aspects with Tipary's DT. See Section \nameref{sec:teaching_free}.

\begin{table}[ht]
\caption{Models in Tipary's robot cell DT \cite{Tipary2022AssemblyNetworks}}
\label{tab:dt_tipary}
\begin{tabular}{p{0.2\linewidth} p{0.7\linewidth} }
\textbf{Model}  & \textbf{Content}
\\ \hline
Kinematic model & Geometry, kinematic behavior and component relations
\\ \hline
Grasp model & Workpiece — manipulator relation through the gripper
\\ \hline
Path model & Manipulation sequence and manipulator motion
\\ \hline
Servo model & Condition-based manipulation (observation-based)
\\ \hline
Metrology model & Providing information to other models and planner tools (observation-based)
\\ \hline
Tolerance model & Represent the tolerance stack-up of the operation to assess its feasibility
\\ \hline
\end{tabular}
\end{table}

\section*{Teaching-free operation with the LAPP digital twin}  \label{sec:teaching_free}

\begin{figure}[ht]
\centering
    \includegraphics[width=0.7\linewidth]{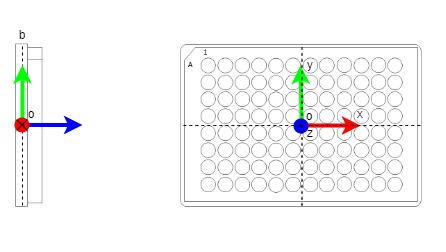}
    \caption{Definition of the plate CS (Overlaid on \cite{ANSI/SLAS2004ANSIPositions}) (US projection). Viewing from an upper view, the origin of the CS (o) is placed to the centerpoint of the plate, whereas the height is set to the middle of the bottom rim (b). When looking at the plate from above with well A1 being on the top left, the x axis points to the right, the y upwards, and z outwards. The TCP for the gripper and the CS of the hand-over site are defined in a way that their origins align with the plate CS and the axes either align or are rotated 90 degrees along the z axis (e.g., in portrait mode gripping, as seen in \ref{fig:gripper_cs}). The site CS is defined similarly, derived from the plate CS.}
    \label{fig:plate_cs} 
\end{figure}

As outlined in \cite{Wolf2021TowardsFramework}, the LAPP framework aims to make the whole set-up process of laboratory robots fully automatic — all the way from map generation with simultaneous localization and mapping (SLAM) to determining the above-mentioned coordinate frame transformations. According to the LAPP approach, the positions (s, ss, sd in Fig. \ref{fig:cs_mobile}) must be defined in relation to the marker frame (m in Fig. \ref{fig:cs_mobile}) in cartesian space. To make this possible, first the workpiece CS must be defined. Since the most common workpiece in this context is an ANSI/SLAS-conform microplate \cite{ANSI/SLAS2004ANSIPositions}, this is taken as a basis. In Fig. \ref{fig:plate_cs}, the axes are derived, as overlaid on a schematically drawing, which is based on the ANSI/SLAS standard \cite{ANSI/SLAS2004ANSIPositions}. 

\begin{figure}[ht]
\centering
    \includegraphics[width=0.7\linewidth]{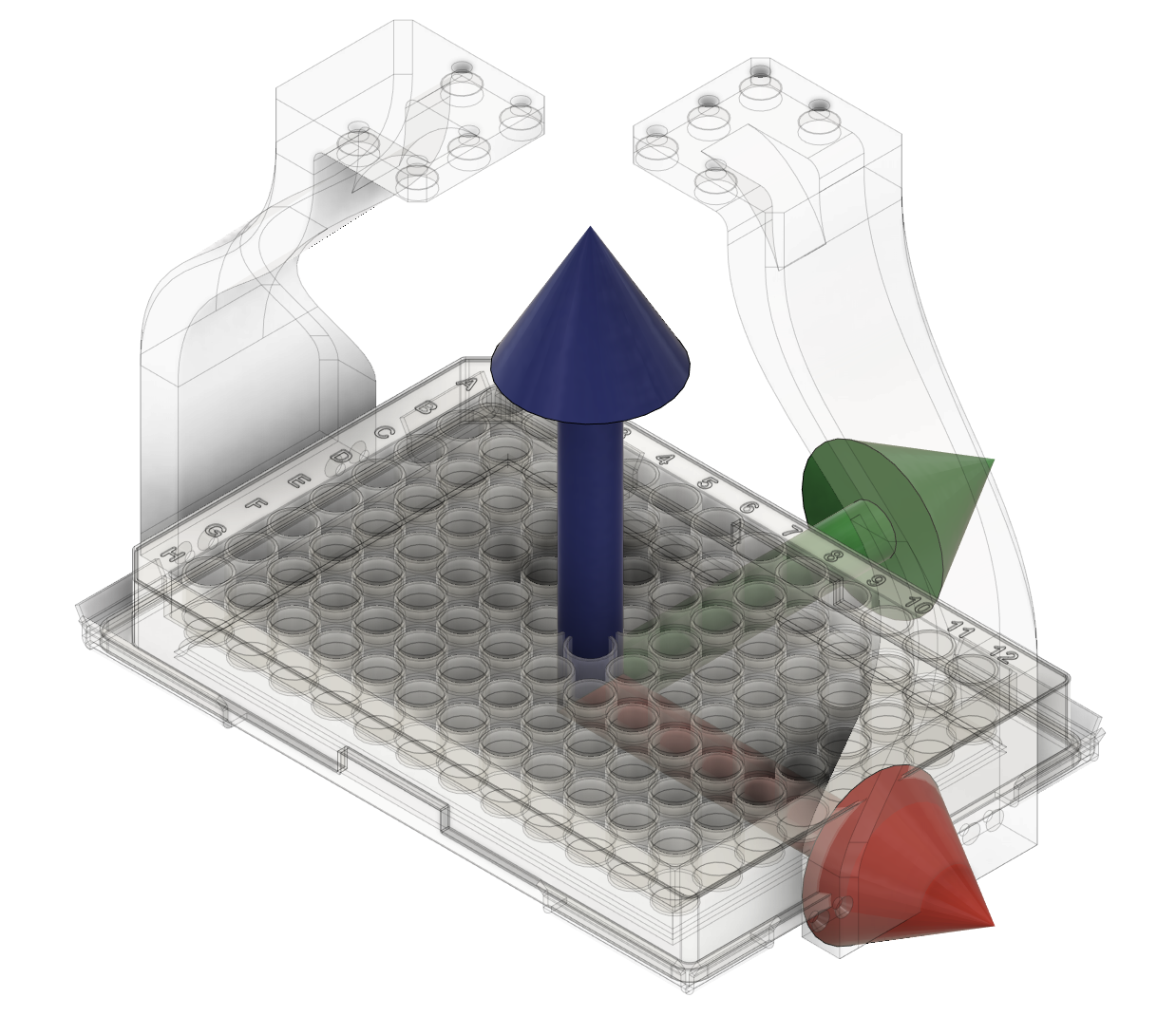}
    \caption{The plate- gripper- and site coordinate systems align in landscape mode gripping}
    \label{fig:gripper_cs} 
\end{figure}

As discussed above, in the LAPP framework the positions are defined in relation to the marker frame in cartesian space and not in the robot's CS in joint space. This enables a robot-independent implementation of the movements thanks to the fact that the positions belong to a certain site of a workstation or device and not to one specific robot. This makes sharing the positions between multiple robots possible. Ultimately, it can be assumed that a certain model of a laboratory device always has the same geometry, i.e., the hand-over site is at a pre-defined location of the device. Supposing that the marker is also at a fixed and known position, the transformation T5 (as displayed in Fig. \ref{fig:cs_mobile} can be deduced. The LAPP framework thus requires the device vendor to include the marker in the design of the device. The marker must be in a visible location, on the same side from where the robot must access it. To ensure compatibility, a specific marker type and size must be specified. The currently available laboratory MoMas \cite{BioseroLab, Kleine-Wechelmann2022DesigningLaboratoryb, RoboticProjects} use different types of ArUco \cite{Garrido-Jurado2014AutomaticOcclusion}, AprilTag \cite{Sagitov2017ARTagRotation} and PI-Tag \cite{Bergamasco2013Pi-Tag:Invariants} markers of different sizes. Assessing these options and selecting a candidate is subject to future work.

It is also the vendor's responsibility to provide the information shown in Table \ref{tab:device_dt} as part of the digital twin prototype of the device (See Section \nameref{sec:dt}, and Table 2 in \cite{Wolf2021TowardsFramework}). Although, only in an ideal world would all vendors commit to make their devices LAPP-compliant. Especially for the launching and ramp-up period, the database will rely more on crowd-sourcing. For this, the DT prototype taxonomy must be provided as an open repository, with clearly defined contribution and admission mechanisms. Also, individual deviations in the device geometry may occur, thus the possibility for adjusting the DT instance by the means of calibration must also be kept open. See Section \ref{sec:challenges}.

\begin{table}[ht]
\centering
\caption{Parameters of the device's digital twin prototype. CS markings defined in \ref{fig:coord_sys}}
\label{tab:device_dt}
\begin{tabularx}{\textwidth}{X p{0.2\textwidth}}
  \makecell{\\ \\ \textbf{Piece of information}} & \makecell{\textbf{As digital twin} \\
  \textbf{parameter} \\
    \textbf{(See Table \ref{tab:dt_params})}} 
  \\
\hline
Robotic tasks, subtasks and motion sequences: e.g., transport by pick \& place
 &
Functionality\\
\hline
Type of workpiece: ANSI/SLAS microplate, Falcon tube etc.
 &
Functionality\\
\hline
Location of the marker (m) in relation to the device CS (d)
 &
Form\\
\hline
Simplified collision geometry of the device, defined in d
 &
Form\\
\hline
Positions. e.g., Hand-over site s, Site approach ss, Device approach sd
 &
Form\\
\hline
Precision requirements
&
Miscellaneous \\
\hline
\end{tabularx}
\end{table}

\pagebreak

In general, DT parameters are distinguished in the LAPP framework as follows:
\begin{itemize}
    \item \textbf{Prototype parameters} are unified for all specimens of the same make and model. See Table \ref{tab:device_dt}
    \item \textbf{Instance parameters} override and extend the prototype parameters with instance-specific data
    \begin{itemize}
        \item \textbf{Static parameters} remain unchanged during the entire lifecycle of the instance. E.g., serial number
        \item \textbf{Volatile parameters} represent the current state of the instance
        \begin{itemize}
            \item \textbf{Observable properties} represent real-time or semi real-time data streams. E.g., sensor streams, continuous progress updates
            \item \textbf{Unobservable properties} are updated only upon specific request. E.g., calibrated positions
        \end{itemize}
    \end{itemize}
\end{itemize}

With the help of this information, a LAPP-enabled MoMa with a calibrated camera system will be capable of performing the setup fully autonomously, as presented in Fig. 1 of \cite{Wolf2021TowardsFramework}. After that, in ideal circumstances (i.e., with calibrated geometries), the pre-defined robotic actions can be performed right away, without requiring any manual input from the integrator or operator. This is the functionality referred to as plug \& play.
As a part of the sequence presented in \cite{Wolf2021TowardsFramework}, the coordinate transformations are determined and the digital twin instances are parameterized, as presented in Table \ref{tab:pp_setup}.

\begin{table}[]
\centering
\caption{The plug \& play setup sequence, markings of Fig. \ref{fig:cs_mobile} used}
\label{tab:pp_setup}
\begin{tabular}{p{0.35\linewidth} p{0.2\linewidth} p{0.15\linewidth} p{0.15\linewidth} }
\textbf{Description} & \textbf{Transform.} & \makecell[l]{\textbf{DT}\\ \textbf{parameter}\\ \textbf{type}} & \makecell[l]{\textbf{DT}\\ \textbf{instance}} \\ \hline 
During the autonomous room discovery procedure, the map is generated
& w & Form & \makecell[l]{Room\\  (instance)}\\ \hline
Simultaneously, the approximate device positions are detected with the markers. Since d and m are already connected by the DT prototype of the device, and the robot is at the PoI d is now defined in w.
& \(T_6(w \rightarrow d)\) & Location & \makecell[l]{Device\\ (instance)}\\ \hline
The hand-over site position is taken from the DT prototype of the device
& \(T_4(m \rightarrow s)\) & Form & \makecell[l]{Device\\  (prototype)}\\ \hline
If necessary, this position can be overridden (calibrated) and stored in the DT instance of the device.
& \(T_4,cal(m \rightarrow s)\) & Form & \makecell[l]{Device\\  (instance)}\\ \hline
The robot kinematics can be re-calibrated and stored in the DT instance of the robot
& \(r \rightarrow g\) & Form & \makecell[l]{Robot\\  (instance)}\\ \hline
\end{tabular}
\end{table}

\begin{figure}[ht]
 \centering
  \subfloat[Global\label{fig:tf_mobile}]{%
        \includegraphics[width=0.8\linewidth]{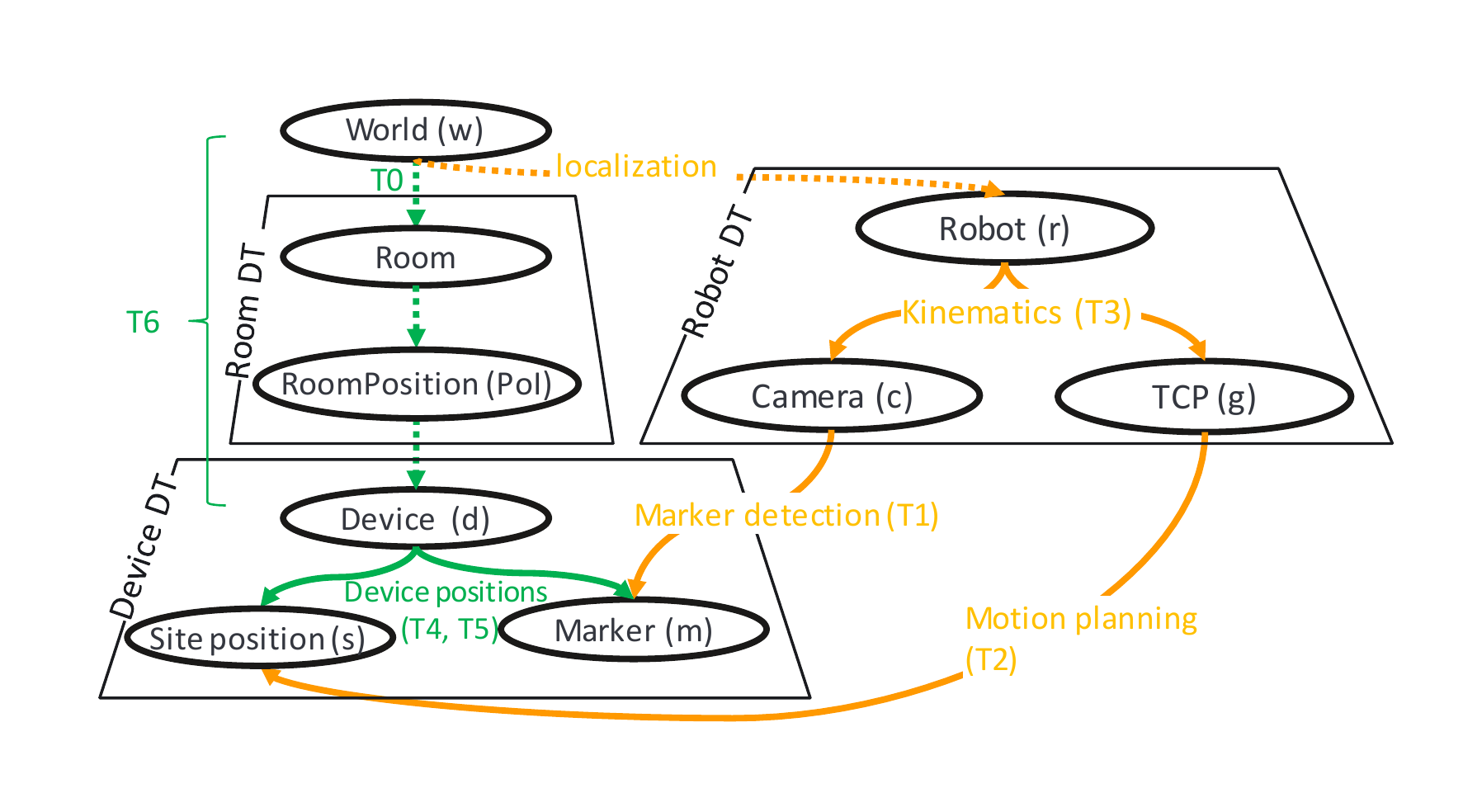}}
    \hspace{5mm}
  \subfloat[Robot-local\label{fig:tf_robot_local}]{%
       \includegraphics[width=0.8\linewidth]{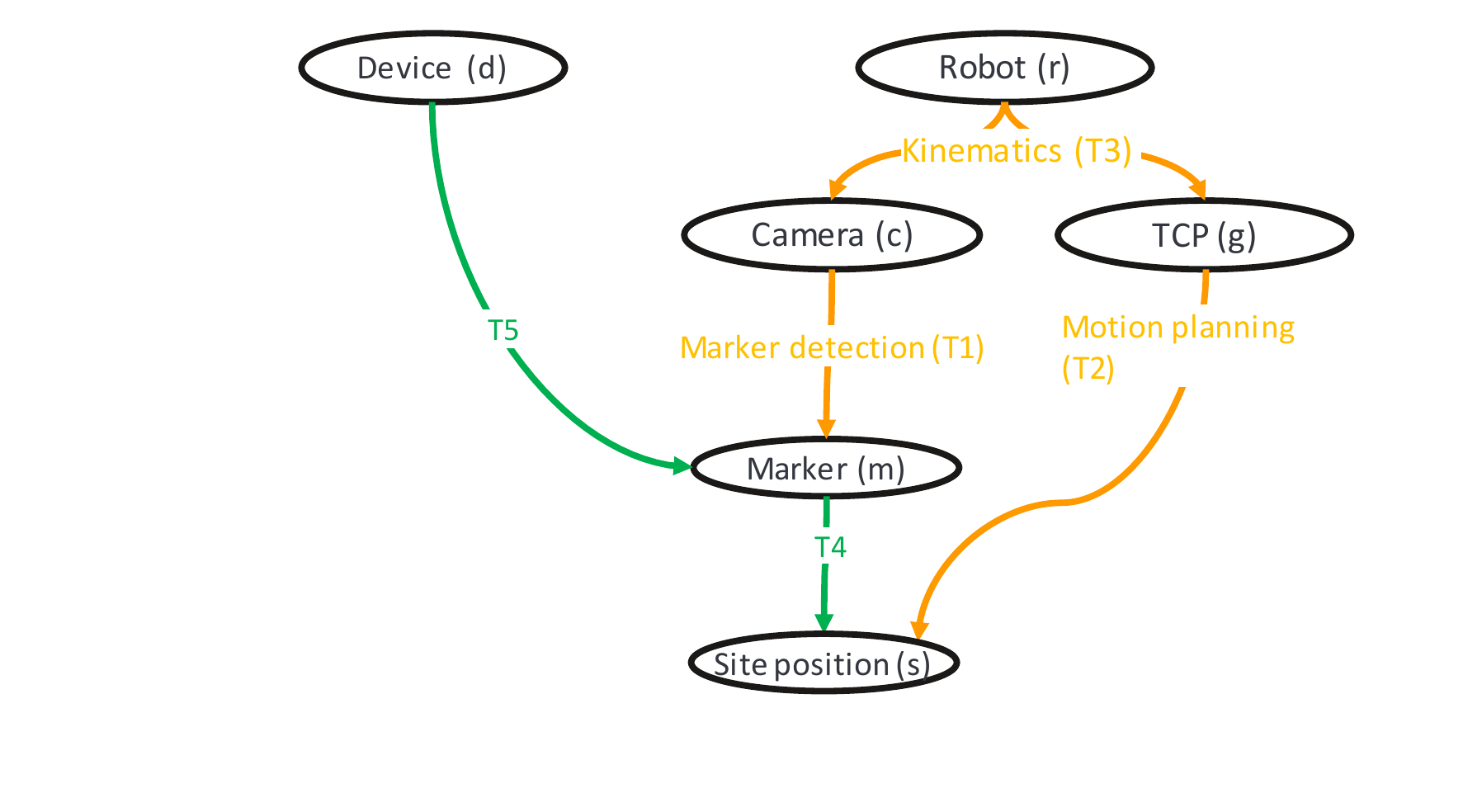}}
    \caption{Coordinate frame transformations (Markings of Fig. \ref{fig:coord_sys} are used.)}
    \label{fig:tf} 
\end{figure}

The relation between the frames and positions in such a system depend on each other in that one frame is defined as a transformation of another frame. This means that the frames can be represented in a graph structure, such as in Fig. \ref{fig:tf}, where the nodes are the frames and the edges are the transformations. The fact that fairly long chains of interlinked frames can form means that the uncertainties of each transformation can accumulate. These uncertainties are inevitably present both in the form of mechanical imperfections causing deviations from ideal geometries and in the inherent inaccuracies of the metrology systems \cite{Thrun2005ProbabilisticSeries}. These must be taken into account by storing the uncertainty for each transformation in the digital twin. Also, a requirement in regard to precision must be specified for the specific robot action. This enables the process controller to match the requirement with the actual precision and determine if the specific system can perform the requested action.

It is also important to mention, that the parent-child relations in a transform tree can change, if a more accurate definition is available. For example, the following case can be considered: When the robot moves to a certain PoI, the system can assume the approximate positions of the devices as defined in relation to the map \(T_6( w \rightarrow d ) \), as shown in Fig. \ref{fig:tf_mobile}. However, the vision system can deliver a far more accurate relation between the robot and the marker ( \(T_1(c  \rightarrow m)\) ), which, along with \(T_4( m \rightarrow s ) \) can be used for motion planning of the pick \& place subtask. For this, a sub-tree from the robot frame (r) can be temporarily created, as presented in Fig. \ref{fig:tf_robot_local}.

\pagebreak

\section*{Challenges and key enablers} \label{sec:challenges}
Traditional laboratory robots are usually taught in joint-space, which means that the robot configuration is stored directly in the form of joint values for each position. This results in the fact that the accuracy of the robot's kinematic model is not of crucial importance. On the contrary, when the positions are specified in world coordinates, the robot controller must perform the inverse kinematics calculations to determine the corresponding joint values. For this, the robot's geometry must be precisely modeled, e.g., by Denavit–Hartenberg parameters or by the Unified Robot Description Format (URDF). These are specified initially for each robot model during the design process. Following the DT notation, these data correspond to the DT prototype of the robot model. Due to manufacturing imperfections, however, each piece of a finished robot has slightly different geometries in reality, which must be readjusted by the means of calibration. These new parameters are stored in the DT instance for the certain robot. To achieve the sub-millimeter precision required for efficient and reliable plate manipulation, the possibility must be kept open for the robot to be re-calibrated.

Another distinction is between online vs. offline teaching. Online teaching means that the positions are taught directly on the physical robot by moving its end-effector to the desired positions manually. In this case, the robot's repeatability is more important than its absolute accuracy; what matters is that the robot is consistent even if systematic error is present. Offline teaching means that the positions are defined without the physical robot, e.g., in the case of the LAPP approach by the device vendor. In this case, the absolute accuracy is just as important as the repeatability, since consistency is not sufficient if the position is off. Therefore, the precision requirements towards a LAPP-enabled robot must be formulated adequately. For digital twin-based offline teaching scenarios in particular, the digital twin closeness to measure such precision is defined by Tipary et al. \cite{Tipary2021GenericTwin} (see Section \nameref{sec:dt}).

Besides kinematics, the mechanics of the robot also play an important role in the robot's precision. One such effect is that no mechanical system is perfectly rigid but flexible and compliant to a certain degree. Reasons behind this are on one hand the mechanical properties of materials: depending on a component's shape and its elastic modulus, it bends under load. On the other hand, the joints of robots consist of bearings and gear systems, which might have mechanical play or backlash. These two effects result in the whole structure diverging from the ideal position when under the effect of internal and external forces, which could be calculated by pure kinematics. These effects must be either mechanically minimized or countered by a suitable controller.

Calibration might also be necessary with regard to the devices' geometries and the positions of the hand-over sites. As described in Section \ref{sec:teaching_free}, this transformation (\(T_4\)) can be fetched from the DT prototype, but overridden (calibrated) on-demand in the DT instance.

\section*{Implementation considerations} \label{sec:technologies}
To make the plug \& play functionality possible, a series of complementary technologies must be consolidated and integrated into a comprehensive stack. To this end, different competing and overlapping solutions were evaluated. Since compatibility is a crucial factor, established and widespread standardized solutions are preferred.

As described in \cite{Wolf2021TowardsFramework}, setting up a robot-based laboratory automation system begins with representing the process on the high-level to enable execution on an all-round process controller. For this purpose, representation formats, such as the Business Process Model and Notation (BPMN) and the Laboratory Open Protocol (LabOP) will be evaluated (See \ref{sec:future_work}. Below this level, the robot-specific actions must be represented in the appropriate abstraction level, for which the LAPP RAPs are proposed and are subject to future work. On the low-level — i.e., on the robot's side — a versatile robot control framework is needed that is capable of incorporating advanced techniques for perception, navigation and motion planning. The Robot Operating System (ROS) proves to fulfill these requirements.

To enable the communication between the different components and levels of the system, a suitable interoperability protocol is needed. The communication protocol developed by the Association Consortium Standardization in Lab Automation (SiLA) provides such a solution specifically for laboratory automation purposes. The organization currently endeavors to extend the robot-related functionality, including the implementation of a SiLA-ROS bridge \cite{SiLA2GitLab} and the unification of the feature definitions \cite{Feature_definitions/ch/unitelabsGitLab}.

Besides that, a suitable platform is needed for the digital twin representation that can incorporate the necessary parameters. The Asset Administration Shell \cite{Details1.0} is identified as a suitable solution. Ontologies for labware, laboratory devices and laboratory robots must be identified, adapted and extended to fulfill the information sharing functionality that is outlined in this paper.

\section*{Future work} \label{sec:future_work}
The LAPP framework, including the LAPP-DT, is at this point a conceptual proposal that is elaborated in the course of an article series entitled \emph{Towards Laboratory Automation Plug \& Play}. This present paper is the second in this series, which takes a step in concretizing the proposal with the focus on robotics. The series follows a top-down approach, which means that the follow-up articles will progress towards the lower technical levels of the framework and provide implementation examples, feasibility studies and comparable benchmarking.

As such, the robotic activity representations (RARs) are identified as the next work package to be elaborated. RARs are considered as one piece of information that must be represented in the digital twin. In general, the digital twin properties must be defined for each type of component in detail, including the identification of suitable standardized representation formats. Further work must focus on the technical implementation by utilizing the technologies discussed in the \nameref{sec:technologies} section. Also, the challenges covered in the section on \nameref{sec:challenges} must be addressed.

The first stage of the implementation is limited to benchtop manipulators and ground-bound mobile manipulators from the hardware side, both with simple parallel grippers. On the use-case side, the scope is focused on pick-and-place type transportation of standardized objects, such as ANSI/SLAS-microplates. Next stages of the development may broaden this scope in regard to both aspects. On one hand, a higher diversity in lab automation robotic hardware can be covered by including emerging technologies. As such, miniature MoMas that drive on the bench or on tracks at ceiling level \cite{ROVERFORMULATRIX} and drones \cite{Kim2018LabSystems} open up new dimensions regarding flexibility and they also require new approaches concerning navigation. On the other hand, advanced end effectors, such as five-finger \cite{PharmaceuticalsCompany} or soft grippers, open up new possibilities to manipulate objects that are fragile and/or have complex geometries. Ultimately, flexible marker-less manipulation can be achieved by the means of advanced perception, such as 3D object detection \cite{Vincze2016LearningHomes} and tactile sensing \cite{Gomes2020GelTip:Manipulation, Thalhammer2021PyraPose:Shift}. These techniques make it possible to detect the pose of the workpiece and to plan the manipulation movement without pre-taught positions. Besides that, the planning also must consider the constraint space containing, inter alia, the obstacles around the working area. Similarly to the other DT parameters, an appropriate standardized format must be identified to store and share this information between the different components of the system.

\section*{Summary} \label{sec:conclusion} 
As a part of the Laboratory Automation Plug \& Play framework, a digital-twin-based approach was proposed that ensures teaching-free setup for laboratory robots. An overview of the state-of-the-art approach showed that current MoMa solutions mostly rely on manually teaching the positions with relation to a fiducial marker. A review of the relevant literature on the digital twin approach was provided to consolidate the nomenclature. The LAPP-DT concept was elaborated on the conceptual level by defining a relative position representation framework as a part of the digital twin of laboratory equipment. After identifying the challenges and the key enablers, specific technological building blocks were formulated, considering the implementation of the framework. A roadmap was presented for the future steps in concretizing the conceptual framework and providing proof-of-concept implementations.

\section*{Conflict of interest statement}
Ádám Wolf and Stefan Romeder-Finger are
employees of Baxalta Innovations GmbH, a Takeda company, Vienna, Austria.
Stefan Romeder-Finger is a stockholder in Takeda Pharmaceutical Company Limited.

\section*{Acknowledgements}
This work was funded by Baxalta Innovations GmbH, a Takeda company.

This work was supported by the Doctoral School of Applied Informatics and Applied Mathematics, Óbuda University.

Péter Galambos and Károly Széll thankfully acknowledge the financial support of this work by the project no. 2019-1.3.1-KK-2019-00007 implemented with the support provided from the National Research, Development and Innovation Fund of Hungary, financed under the 2019-1.3.1-KK funding scheme. Péter Galambos is a Bolyai Fellow of the Hungarian Academy of Sciences. 
 
 \pagebreak



\bibliography{references.bib}



%
%
%


\end{document}